\documentclass[sigconf,screen]{acmart}
\setcopyright{none}
\usepackage{microtype}
\usepackage{graphicx}
\usepackage{subfigure}
\usepackage{booktabs} 

\usepackage{url}
\usepackage[utf8]{inputenc} 
\usepackage[T1]{fontenc}    
\usepackage{url}            
\usepackage{booktabs}       
\usepackage{amsfonts}       
\usepackage{nicefrac}       
\usepackage{microtype}      
\usepackage{amsmath}
\usepackage{graphicx}
\usepackage{epigraph}
\usepackage[export]{adjustbox}
\usepackage[final]{pdfpages}
\usepackage{mathrsfs}
\usepackage{makecell}
\usepackage{enumitem}
\usepackage{pdfpages}
\usepackage{longtable}
\usepackage{multirow}
\usepackage{tcolorbox}
\usepackage{tabu}
\usepackage{algorithm}
\usepackage{xcolor}
\usepackage{algpseudocode}
\usepackage{xspace}
\usepackage{bm}
\usepackage{enumitem}
\usepackage{tcolorbox}

\usepackage{listings} 
\definecolor{isarblue}{HTML}{006699}
\definecolor{isargreen}{HTML}{009966}
\lstdefinelanguage{isabelle}{%
    keywords=[1]{type_synonym,datatype,fun,abbreviation,definition,proof,lemma,theorem,corollary},
    keywordstyle=[1]\bfseries\color{isarblue},
    keywords=[2]{where,assumes,shows,and},
    keywordstyle=[2]\bfseries\color{isargreen},
    keywords=[3]{if,then,else,case,of,SOME,let,in,O,qed,next,by,show,thus,hence,have},
    keywordstyle=[3]\color{isarblue},
}
\newcommand{\sysname}{Baldur\xspace}

\usepackage{filecontents}
\usepackage{pgfplots}

\makeatletter
\let\c@figure\c@table
\let\ftype@figure\ftype@table
\let\ext@figure\ext@table

\makeatother

\usepackage{hyperref}

\title[Baldur: Whole-Proof Generation and Repair with Large Language Models]{Baldur: Whole-Proof Generation and Repair\\with Large Language Models}

\begin{document}
	
\author{Emily First}
\orcid{0000-0002-2896-2928}
\affiliation{%
	\institution{University of Massachusetts}
	\city{Amherst}
	\state{MA}
    \postcode{01003-9264}
	\country{USA}}
\email{efirst@cs.umass.edu}

\author{Markus N. Rabe}
\affiliation{%
	\institution{Google, Inc.}
	\state{CA}
	\country{USA}}
\email{mrabe@google.com}

\author{Talia Ringer}
\affiliation{%
	\institution{University of Illinois Urbana-Champaign}
	\state{IL}
	\country{USA}}
\email{tringer@illinois.edu}

\author{Yuriy Brun}
\orcid{0000-0003-3027-7986}
\affiliation{%
	\institution{University of Massachusetts}
	\city{Amherst}
	\state{MA}
    \postcode{01003-9264}
	\country{USA}}
\email{brun@cs.umass.edu}

\begin{abstract}

Formally verifying software properties is a highly desirable but labor-intensive task. 
Recent work has developed methods to automate formal verification using proof assistants, such as Coq and Isabelle/HOL, e.g., by training a model to predict one proof step at a time, and using that model to search through the space of possible proofs.
This paper introduces a new method to automate formal verification: We use large language models, trained on natural language text and code and fine-tuned on proofs, to generate whole proofs for theorems
at once, rather than one step at a time. We combine this proof generation model with a 
fine-tuned repair model to repair generated proofs, further increasing proving power.
As its main contributions, this paper demonstrates for the first time that:
(1)~Whole-proof generation using transformers is possible and is as effective as search-based techniques without requiring costly search.
(2)~Giving the learned model additional context, such as a prior failed proof attempt and the ensuing error message, results in proof repair and further improves automated proof generation.
(3)~We establish a new state of the art for fully automated proof synthesis.
We reify our method in a prototype, \sysname, and evaluate it on a benchmark of 6,336 Isabelle/HOL theorems and their proofs.
In addition to empirically showing the effectiveness of whole-proof generation, repair, and added context, we show that \sysname improves on the state-of-the-art tool, Thor, by automatically generating proofs for an additional 8.7\% of the theorems.  Together, \sysname and Thor can prove 65.7\% of the theorems fully automatically.
This paper paves the way for new research into using large language models for automating formal verification.

\end{abstract}

\maketitle



\pagestyle{plain}
\thispagestyle{plain}

\section{Introduction}
\label{sec:intro}

\looseness-1
Formal software verification --- proving software correctness and other properties --- is one of the most challenging tasks software engineers can undertake.  It is highly effective at producing high quality software.
For example, CompCert, a C compiler verified using the Coq interactive theorem prover~\cite{coq}, was the only compiler on a list including the ubiquitous GCC and LLVM, in which a comprehensive study found no bugs~\cite{Yang11a}.
Similarly, the seL4 project resulted in an highly reliable operating system microkernel~\cite{Klein2009sel4}.
However, the cost of manual formal verification --- writing the proofs --- is often prohibitive.
For example, the proof of the C compiler is more than three times as long as the compiler code itself~\cite{Leroy09}.
As a result, recent research has focused on automated proof synthesis, which can lead to fully automating formal verification.

There are two promising approaches for automating proof synthesis.  The first is to use \emph{hammers}, such as Sledgehammer~\cite{Paulson23} for the Isabelle proof assistant. Hammers iteratively apply known mathematical facts using heuristics. The second is to use search-based \emph{neural theorem provers}, such as DeepHOL~\cite{Bansal2019HOList}, GPT-f~\cite{polu2020gptf}, TacticZero~\cite{Wu2021TacticZero}, Lisa~\cite{Jiang21Lisa}, Evariste~\cite{Lample2022HypertreeProofSearch}, Diva~\citep{first2022diva}, TacTok~\cite{First20oopsla}, and ASTactic~\cite{yang2019learning}.
Given a partial proof and the current \emph{proof state} (which consists of the current goal to prove and the list of known assumptions), these tools use neural networks to predict the next individual \emph{proof step}. They use the \emph{proof assistant} to evaluate the proposed next proof steps, which returns a new set of proof states. 
Neural theorem provers rely on diverse neural architectures, such as Wavenet~\cite{Bansal2019HOList, OordDZSVGKSK16Wavenet}, graph neural networks~\cite{Paliwal2020Graph}, short long-term memory models~\cite{first2022diva}, and language models with the transformer architecture~\cite{polu2020gptf, Han22PACT}.

In this paper, we propose \sysname, a different, simpler approach to proof synthesis.
We show that using large language models (LLMs), fine-tuned on proofs, can produce entire proofs for theorems.
LLMs are scaled-up transformer models trained on a large amount of text data, including natural language and code, that have proven to be remarkably effective across a wide variety of applications, including question answering, and text and code generation~\cite{brown2020gpt3, chowdhery2022palm}.
Here, we show their remarkable effectiveness for whole proof generation.

\begin{tcolorbox}

The main contributions of our work are: 
\begin{itemize}[labelwidth=0.7em, labelsep=0.6em, topsep=-0.5ex, itemsep=0ex, parsep=0ex, leftmargin=1.2em]

\item We develop \sysname, a novel method that generates whole formal proofs using LLMs, without using hammers or computationally expensive search.

\item We define a proof repair task and demonstrate that repairing incorrectly generated proofs with LLMs further improves \sysname's  proving power when the LLM is given access to the proof assistant's error messages.

\item We demonstrate empirically on a large benchmark that Baldur, when combined with prior techniques, significantly improves the state of the art for theorem proving.

\end{itemize}

\end{tcolorbox}

We design \sysname to be able to work with any LLM internally, but we evaluate our implementation using two versions of Minerva~\cite{Lewkowycz2022Minerva}, one with 8 billion parameters and another with 62 billion parameters.
By contrast, existing tools that use (L)LMs for theorem proving, either predict individual proof steps~\cite{Han22PACT, Jiang21Lisa, Jiang2022Thor}, or rely on few-shot prompting and require the existence of natural language proofs as hints~\cite{jiang2022draftsketchprove}.

We evaluate \sysname on the PISA dataset~\citep{Jiang21Lisa} of Isabelle/HOL theorems and their proofs used in recent state-of-the-art Isabelle/HOL proof synthesis evaluations~\cite{Jiang21Lisa, Jiang2022Thor}.  The dataset consists of 183K theorems, of which we use 6,336 for measuring effectiveness. Our evaluation answers the following research questions: 
\begin{itemize}

  \item[RQ1:] How effective are LLMs at generating whole proofs? \\
  \textbf{LLMs outperform small-model-driven search-based methods.} \sysname (without repair) is able to generate whole proofs for 47.9\% of the theorems completely automatically, whereas search-based approaches prove 39.0\%~\cite{Jiang2022Thor}.

  \item[RQ2:] Can LLMs be used to repair proofs? \\
  \textbf{LLMs can repair proofs, including their own erroneous proof attempts.} \sysname proves an additional 1.5\% of the theorems when given access to a previous erroneous proof attempt and the error messages produced by the proof assistant, even when controlling for the computational cost of the additional inference.
  The error message is crucial for this improvement.

  \item[RQ3:] Can LLMs benefit from using the context of the theorem? \\
  \textbf{In-context learning is remarkably effective for LLM-based theorem proving.} With context, \sysname proves 47.5\% of the theorems, but only 40.7\% without context for the same model size.

  \item[RQ4:] Does the size of the LLM affect proof synthesis effectiveness? \\
  \textbf{Larger LLMs do perform better,} suggesting that our approach will continue to improve with further developments in LLM research.

  \item[RQ5:] How do LLMs compare to other state-of-the-art proof generation methods? \\
  {\bf\sysname complements state-of-the-art approaches by proving theorems they do not.}
  Together with Thor~\cite{Jiang2022Thor}, a tool that combines a learned model, search, and a hammer, \sysname can prove 65.7\% of the theorems, whereas Thor alone proves 57.0\%.
  These findings suggest that LLM- and search-based methods' ideas complement each other and can work together to further improve the automation of formal verification.
  An ensemble of 10 different fine-tuned \sysname models proves 58.0\%.

\end{itemize}

By leveraging LLMs, \sysname simplifies the proof synthesis pipeline, greatly reducing the complexity and cost of the fine-grained interaction between the prediction model and the proof assistant that search-based methods require. This reduction enables us to leverage the power of LLMs, which would be prohibitively computationally expensive if synthesis required as many LLM queries as search-based methods. Further, those calls would require re-encoding with each step the additional information the LLM might need, whereas our approach allows us to make a single call and process the context only once, sampling multiple proofs of multiple proof steps, at once.\footnote{Alternatively path advanced caching strategies in the prediction servers of large language models could address this problem. This is beyond the scope of our work.}  Overall, our study strongly suggest that LLMs are a very promising direction of research for automating formal verification, and identifies several new avenues for future explorations.

\section{The \sysname Approach}
\label{sec:method}

Prior approaches to proof synthesis employ a neural model to predict the 
next \emph{proof step} given the current \emph{proof state}. 
The proof step predictions then guide a search 
strategy, such as best-first search or depth-first search. Throughout the search, the proof 
assistant needs to check each proof step prediction to determine whether it is valid.
This means that existing proof synthesis tools require a tight interaction 
between the neural network and the proof assistant.
As we move to using LLMs, this results in complex systems, as LLMs need to run on specialized hardware (GPUs or TPUs), while proof assistants run on CPUs.

We explore a simpler, yet effective method: fine-tuning LLMs to generate complete proofs.
This simplification avoids the fine-grained interaction between neural model and the 
proof assistant, allowing us to run the jobs of generating proofs and checking completely separately.
Besides reducing complexity, this can also improve efficiency, because
(1)~it enables us to use large batch sizes, which can significantly improve hardware utilization during inference (cf.~\cite{Pope2022Inference}), and (2)~when providing additional context to the model, the context now does not have to be reprocessed for each proof step, but only once per proof.

We fine-tune LLMs on proof data to generate entire proofs and explore the impact of 
giving the LLMs additional information. Our approach and implementation include the following:
\begin{itemize}[labelwidth=0.7em, labelsep=0.6em, topsep=-0.5ex, itemsep=0ex, parsep=0ex, leftmargin=1.2em]

    \item We fine-tune an LLM to generate an entire proof given only the theorem statement. We call this model the \emph{proof generation model} (Section~\ref{ssec:generation}).

    \item We provide a model a proof attempt that did not check along with the corresponding \emph{error message} from the proof assistant so that the model may attempt to find a better proof. We call this model the \emph{proof repair model} (Section~\ref{ssec:repair}).

    \item We provide text from the same \emph{theory file} that the problem was taken from.
    We add only the lines from the theory file that immediately precede the theorem we want to prove. We call this added information the \emph{theory file context} and we add it to the 
    proof generation model (Section~\ref{ssec:context}).

    \item The LLM that we fine-tune at the core of all of this is Minerva~\citep{Lewkowycz2022Minerva}, which is pretrained on a mathematics corpus. We describe our \sysname-specific implementation details for how we use this model (Section~\ref{ssec:model}).
\end{itemize}

These fine-tuned LLMs and their interaction with the Isabelle proof assistant make up our tool \sysname. This section details the \sysname approach, which includes creating training datasets and leveraging LLMs to generate and repair proofs.

\subsection{Proof Generation}
\label{ssec:generation}

\begin{figure}
    \centering
    \includegraphics[scale=0.9]{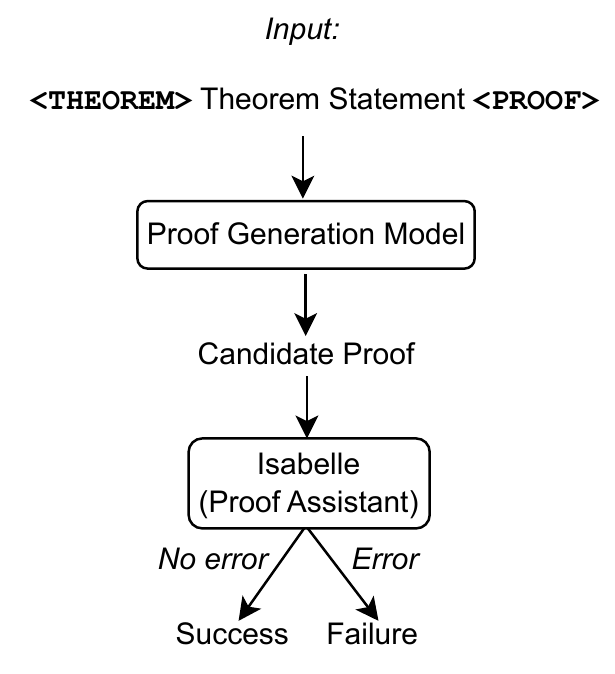}
    \vspace{-2ex}
    \caption{An example of using the proof generation model to generate a proof.}
    \label{fig:generation_process_example}
\end{figure}

Existing proof generation methods using neural models generate the proof one step at a time. In contrast, our approach generates the entire proof, as illustrated with a single example in Figure~\ref{fig:generation_process_example}. We use only the theorem statement as input to our \emph{proof generation model}. We then sample a proof attempt from this model and perform 
proof checking using Isabelle. If Isabelle accepts the proof attempt without an error, then we have proven the theorem. Otherwise, we can try sampling another proof attempt from the 
proof generation model. Explicitly, the input and output of our proof generation model is as follows:

\begin{itemize}[labelwidth=0.7em, labelsep=0.6em, topsep=-0.5ex, itemsep=0ex, parsep=0ex, leftmargin=1.2em]

    \item \textbf{Input:} theorem statement.

    \item \textbf{Output:} candidate proof.

\end{itemize}

\paragraph{Example.}

To illustrate the power of the proof generation approach in our tool \sysname, we first consider, as an example, the theorem \lstinline{fun_sum_commute}.

\begin{lstlisting}
lemma fun_sum_commute:
  assumes "f 0 = 0" and "&$\land$&x y. f (x + y) = f x + f y"
  shows "f (sum g A) = (&$\Sigma$&a&$\in$&A. f (g a))"   
\end{lstlisting}

The theorem states that for an additive function $f$ where $f(0)=0$, and an arbitrary function $g$, applying $f$ on the sum of the set resulting from applying $g$ on each element in a given set is equal to the sum of applying $g$ followed by $f$ to each element in that set.
This theorem is from a project in the Archive of Formal Proofs called Polynomials, specifically in the file \texttt{Utils.thy}.

The human-written proof distinguishes between two cases: when the set is finite and when it is not.
Induction is used for the finite set case.

\begin{lstlisting}
proof (cases "finite A")
  case True
    thus ?thesis
    proof (induct A)
      case empty
      thus ?case by (simp add: assms(1))
    next
      case step: (insert a A)
      show ?case by (simp add: 
        sum.insert[OF step(1) step(2)]
        assms(2)
        step(3))
    qed
  next
    case False
      thus ?thesis by (simp add: assms(1))
qed
\end{lstlisting}

If we were to derive a training example from this example, the input would be theorem statement and the target would be this human-written proof. 

Our tool \sysname, using the proof
generation model, is able to generate the following correct proof for this statement. 

\begin{lstlisting}
  by (induct A rule: infinite_finite_induct) 
  (simp_all add: assms)  
\end{lstlisting}

\sysname recognizes that induction is necessary and applies a special induction rule called \lstinline{infinite_finite_induct}, following the same overarching approach as the human-written proof, but much more succinctly. It is interesting to note that Sledgehammer, the hammer for Isabelle, cannot prove this theorem by default, as it requires induction.


\paragraph{Training Data Creation.} 

To train the proof generation model, we construct a new proof generation dataset. 
Existing datasets for training models in neural theorem provers contain examples of individual proof steps. Each training example includes, at minimum, the proof state (the input) and the next proof step to apply (the target).  
Given a dataset that contains individual proof steps, we want to create a new dataset so that we can train models to predict entire proofs at once.
So we extract the proof steps of each theorem from the dataset and concatenate them to reconstruct the original proofs.
We use this data to generate training examples for the proof generation model, where the input consists of the theorem statement and the target consists of the proof.

In particular, this means that we drop the \emph{proof states} from the dataset, which make up most of the text in the dataset.
We argue that for Isabelle proofs this is not necessarily a problem, as Isabelle uses a declarative proof language that is designed to be human-readable.
This is in contrast to other proof assistants, such as Coq, where the proofs are typically written in a procedural style that is not easy to interpret for humans without using the proof assistant to generate the intermediate proof states.

\paragraph{Inference.}
We fine-tune an LLM on our data to predict the entire proof given only a theorem statement.
To synthesize a proof using the fine-tuned LLM, we provide a potentially unseen theorem statement and sample a fixed number of sequences (typically 16 or 64) from the language model.
We tune the sampling temperature from a small set (between 0.0 and 1.4 in increments of 0.2), which is a multiplicative factor on the log probabilities of the distribution of tokens sampled in each step.

\paragraph{Proof checking.}
After sampling proofs from the model, we check all of them with the proof assistant.
This means that we first load the context in which the theorem was originally proven and then replace the original proof of the theorem with the one we sampled from the model.
If Isabelle accepts any of the sampled proofs, we report the theorem as proven.

\subsection{Proof Repair}
\label{ssec:repair}

\begin{figure}[t]
    \centering
    \includegraphics[scale=0.9]{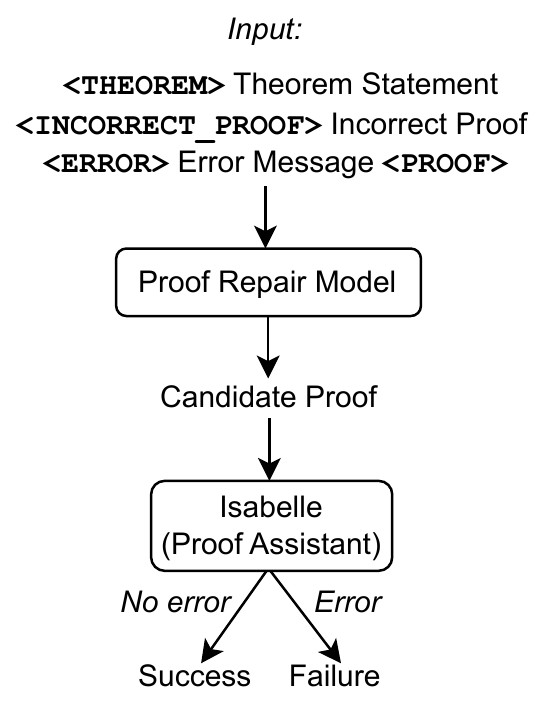}
    \caption{An example of using the proof repair model to repair an incorrect proof.}
    \label{fig:repair_process_example}
\end{figure}

If a proof is not accepted, Isabelle returns an error message that is intended to help humans with debugging their proof script.
Existing proof generation methods, however, have no way to leverage error messages.

Building off our proof generation approach, we explore the use of error messages to improve neural theorem provers by developing a proof repair approach.
Starting with just the problem statement, we apply the proof generation model from Section~\ref{ssec:generation} to sample a proof attempt.
If Isabelle accepts the proof attempt, we can stop.
Otherwise, we use the error message returned by the proof checker and the incorrect proof
attempt to construct an example to serve as input to the \emph{proof repair model}. As depicted in Figure~\ref{fig:repair_process_example}, we use the theorem statement, the incorrect proof, and the error message as input to our proof repair model. We then sample the proof attempt 
from this model, and perform proof checking in the same way as the proof generation approach. Explicitly, the input and output of our proof repair approach 
pipeline are as follows:

\begin{itemize}
    \item \textbf{Input:} theorem statement, incorrect proof, error message.
    \item \textbf{Output:} candidate proof.
\end{itemize}

\paragraph{Example}
Starting from the theorem \lstinline{fun_sum_commute}, we illustrate an example of the proof repair approach in our tool \sysname. We apply the proof generation model to obtain more proof attempts. The following is a proof attempt generated by \sysname, which fails in the proof checker.

\begin{lstlisting}
  proof (induct A)
  case (insert x A)
  thus ?case
    by (simp add: assms(2))
  qed simp
\end{lstlisting}

\begin{figure}[t]
    \centering
    \includegraphics[scale=0.75]{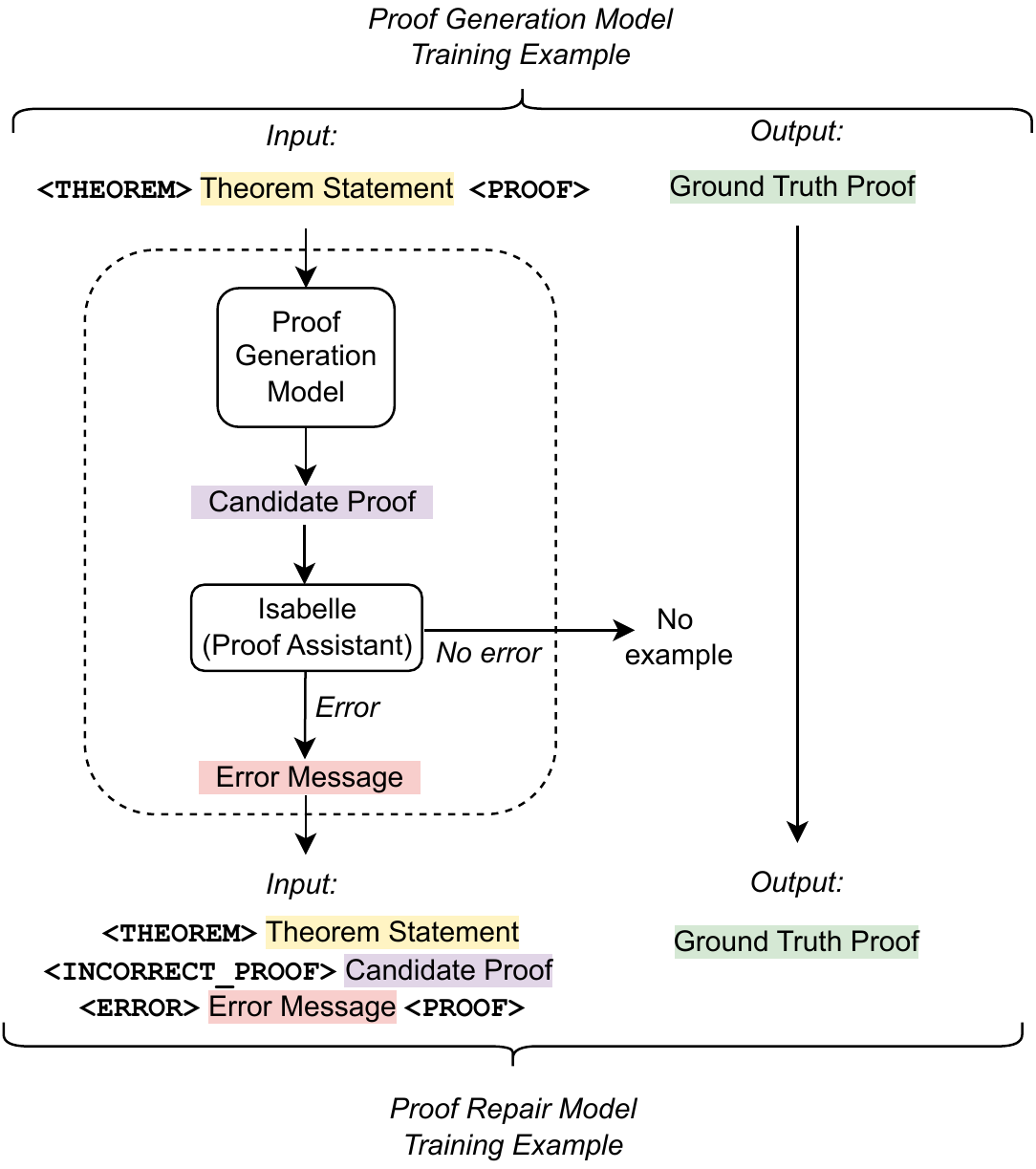}
    \caption{Training data creation for the proof repair model.}
    \label{fig:repair_model_data_creation}
\end{figure}

\sysname attempts to apply an induction, but fails to first break down the proof into two cases (finite vs. infinite set).
Isabelle returns the following error message:

\begin{lstlisting}
Step error: Unable to figure out induct rule
At command "proof" (line 1)
\end{lstlisting}

The error message details where the error occurs (line 1) and that the issue is regarding the induct rule. With these strings as input, using the proof repair model, \sysname can attempt to generate a correct proof for this statement. If we want to instead derive a proof repair training example from these strings, we concatenate the theorem statement, the failed proof attempt, and the error message to serve as the input, and we use the correct human-written proof (recall from previous section) as the target.

\paragraph{Training Data Creation.} To train the proof repair model, we need to generate a proof repair training set. Figure~\ref{fig:repair_model_data_creation} details the training data creation process. Using the proof generation model, we sample one proof with temperature 0 for each problem in the original training set used to train the proof generation model. Using the proof assistant, we record all failed proofs and their error messages. We then proceed to construct the new proof repair training set. For each original training example, we concatenate the theorem statement, the (incorrect) candidate proof generated by the proof generation model, and the corresponding error message to obtain the input sequence of the new training example. For the target sequence, we reuse the ground truth proof from the original training example. We fine-tune the pretrained LLM on the proof repair training set to obtain the proof repair model.

\subsection{Adding Context}
\label{ssec:context}

LLMs possess impressive in-context learning abilities (cf.~\cite{brown2020gpt3,chowdhery2022palm}) that allow them to flexibly use information that is provided as part of the input sequence (and, in fact, as part of their own output~\cite{Nye2021Scratchpad,Wei2022COT}).
In order to explore to what extent in-context learning can help in the theorem proving domain, we extend their inputs with potentially helpful context. Adding to our proof generation approach, we use the theory file contexts (the lines preceding the theorem statement) as input
to our \emph{proof generation model with context}. Explicitly, the input and output of our proof generation model with context is as follows:

\begin{itemize}
    \item \textbf{Input:} theory file context and theorem statement.
    \item \textbf{Output:} candidate proof.
\end{itemize}

\paragraph{Example.} Continuing the example, the theory file context directly preceding 
\lstinline{fun_sum_commute} is the following theorem statement and 
its associated proof.

\begin{lstlisting}
lemma additive_implies_homogenous:
  assumes "&$\land$&x y. f (x + y) = f x + 
  ((f (y::'a::monoid_add))::'b::cancel_comm_monoid_add)"
  shows "f 0 = 0"
proof -
  have "f (0 + 0) = f 0 + f 0" by (rule assms)
  hence "f 0 = f 0 + f 0" by simp
  thus "f 0 = 0" by simp
qed
\end{lstlisting}

The proof generation model with context in \sysname can leverage this additional information. Strings that appear in the theorem statement for \lstinline{fun_sum_commute}, such as 
\lstinline{"f 0 = 0"}, appear again in this context, and so the additional information 
surrounding them could help the model make better predictions.

\paragraph{Training Data Creation.}
We add the lines of the theory file that precede the theorem statement to serve as additional 
context. This means that context can include statements, such as the previous theorems, definitions, proofs, and even natural language comments.
To make use of the available input length of LLMs, we first add up to 50 preceding statements from the same theory file. 
During training, we first tokenize all these statements, and then we truncate the left of the sequence to fit the input length.

\paragraph{Premise Selection}
Many proofs make frequent use of definitions and previously proven statements, also known as \emph{premises}.
Some neural theorem provers, such as HOList~\cite{Bansal2019HOList}, focus entirely on the problem of selecting the right set of premises, which has been shown to be quite successful in theorem proving.

Premise selection is clearly similar to the addition of context in some aspects, but we want to emphasize some key differences: (1) Adding context is an extremely simple technique that only requires rudimentary text processing, (2) by adding the preceding lines of the theory file, the model can only observe a small fraction of the available premises, (3) most of the added context consists of proofs.



\subsection{Large Language Model}
\label{ssec:model}

We use Minerva~\citep{Lewkowycz2022Minerva}, a large language model pretrained on a mathematics corpus based on the PaLM~\cite{chowdhery2022palm} large language model.
Specifically, we use the 8 billion parameter model and the 62 billion parameter model. 
The Minerva architecture follows the original Transformer architecture~\cite{vaswani2017attention}, but has some noteworthy differences.
It is a decoder-only transformer with maximum sequence length of 2,048 tokens. The model uses
\begin{itemize}

  \item rotary position encodings~\cite{Su2021Rotary} instead of sinusoidal absolute position embeddings,

  \item parallel layers~\cite{black-etal-2022-gpt}, which compute the feed forward layer and the attention layer in parallel and add up their results instead of computing them in sequence, and

  \item multi-query attention, which uses a single key-value pair per token per layer for faster decoding~\cite{Shazeer2019multiquery}.

\end{itemize}
As this model is not a contribution of this paper, we refer the reader to prior work for lower-level details on the Minerva architecture~\cite{chowdhery2022palm}.

\paragraph{\sysname-specific implementation details}
The proof generation task naturally consists of an input, which is the theorem statement (potentially augmented with additional information), and the output (target), which is the proof for the theorem.
To work with the decoder-only model, we concatenate the inputs and targets, but the loss is only computed over the target during fine-tuning.
The inputs use bidirectional attention while the targets use causal attention as in PrefixLM~\cite{Raffel2020PrefixLM}.

As the transformer has a maximum context length of 2048, we pad the sequences with zeros if they are too short, and we need to truncate them if they are too long.
Inputs to the model are truncated to the maximum input length by dropping tokens on the left.
The rationale for dropping tokens on the left is that the additional context is given before the theorem statement, and can be truncated more safely than the theorem statement itself.
Similarly, targets (i.e. the proof to generate) are truncated on the right to the maximum target length.

We used a maximum input length of 1536 and a maximum target length of 512 all experiments but the repair study, which used 1024 and 1024 instead.
We use a drop-out rate of 0.1 for both generation and repair models to address overfitting.

During sampling from the language model we restrict the choice of the next token to the 40 tokens with the highest score, also called top-K sampling~\cite{Fan2018Hierarchical}.
We sample sequences with a maximal length of 256 tokens. The model was trained to generate up to 512 tokens, but since most successful proofs are relatively short, this limitation has little impact on the proof rate while saving some compute.

We use a batch size of 32, and fine-tune for up to 100,000 steps, but we observed that the model begins to overfit to the training set after 50,000 to 70,000 steps.
For inference, we selected checkpoints from just before the model started to overfit.

\section{Evaluation}
\label{sec:experiments}

In this section we present several experiments and discuss the following research questions:

\begin{itemize}

  \item[RQ1:] How effective are LLMs at generating whole proofs?

  \item[RQ2:] Can LLMs be used to repair proofs?

  \item[RQ3:] Can LLMs benefit from using the context of the theorem?

  \item[RQ4:] Does the size of the LLM affect proof synthesis effectiveness?

  \item[RQ5:] How do LLMs compare to other SOTA proof generation methods?

\end{itemize}
To answer these questions, we trained several language models using the approach from
Section~\ref{sec:method}, and evaluated them on the PISA benchmark (see Section~\ref{ssec:pisa}).
Our main results can be found in Table~\ref{tab:results} and in Figure~\ref{fig:blueplot}.

\subsection{Experimental Setup}

\begin{table}
    \centering
    \begin{tabular}{lcc}
    \hline
        Model              & 16 samples & 64 samples \\\hline
        Baldur 8b generate & 34.8\%\phantom{$^*$} & 40.7\% \\
        Baldur 8b generate + repair & 36.3\%$^*$ & --- \\
        Baldur 8b w/ context & 40.9\%\phantom{$^*$} & 47.5\% \\
        Baldur 62b w/ context & 42.2\%\phantom{$^*$} & 47.9\% \\\hline
        Baldur 8b w/ context $\cup$ Thor & --- & 65.7\%
    \end{tabular}
    \vspace{2mm}
    \caption{Proof rate of different models. \\ 
    $^*$The repair approach uses half the number of samples, and then one repair attempt for each sample.}
    \label{tab:results}
\end{table}

\paragraph{Machine specification}
For most of the training runs of the 8b model, we used 64 TPUv3 cores distributed across 8 hosts.
For training the 62b model, we used 256 TPUv3 cores distributed across 32 hosts.
For most inference jobs, we used between 32 inference servers using 8 TPUv3 cores each.


\paragraph{Proof Checker}
We use the PISA codebase~\citep{Jiang21Lisa} under a BSD 3-clause 
license, which allows us to interact with the Isabelle proof assistant to check proofs.
To run large jobs of the proof checker, we package it in a Docker container and run it on GCP.
We extended the proof checker to discard any proofs that contain ``sorry'' or ``oops'', which are keywords that skip proofs, but otherwise pass the proof checker.
We apply a timeout of 10 seconds to each proof step in the proof checker.

\subsection{PISA Benchmark}
\label{ssec:pisa}
We derive our datasets from the PISA dataset~\citep{Jiang21Lisa}, which includes
the Isabelle/HOL repository under a BSD-style license and the Archive of
Formal Proofs (AFP) from October 2021. The AFP is a
large collection of Isabelle/HOL proof developments.
PISA includes the core higher-order logic library of Isabelle, as well as
a diverse library of proofs formalised with Isabelle. This includes
mathematics proofs and verification of software and hardware systems.
The PISA dataset comes with a 95\%/1\%/4\% split of theorems for
the training/validation/test sets, which we follow in this work as well. 

For the test set, prior work randomly chose 3,000 theorems from the test set to report their results on.
We report our results on the complete test set.
Some entries in the dataset are not proper theorems (starting with the keyword ``lemmas'' instead of ``lemma''), which we filter out, as did prior work.
This leaves us with a total of 6,336 theorems in our test set (originally 6,633 theorems).

It is worth noting that, as with any LLM-based work, there is the potential for proofs from the test set to have leaked into the LLM pretraining data. While the pretraining data for the Minerva LLM at the base of our models does not include the PISA dataset, it does contain code that may include some Isabelle/HOL proofs found in PISA. This should be kept in mind when interpreting the results.

\subsection{RQ1: How effective are LLMs at generating whole proofs?}
\label{ssec:rq_generate}


We aligned our methodology with the methodology described in Thor~\cite{Jiang2022Thor} to enable a comparison between various methods.
The Thor paper includes informative baselines for the PISA benchmark, including Sledgehammer, a method relying on heuristic search, and a language model approach using search.

Sledgehammer and the search-based language model approach achieve 25.6\% and 39.0\%, respectively.
In comparison, our naive proof generation approach with an 8b language model achieves a proof rate of 34.8\% with 16 samples and of 40.7\% with 64 samples.
The comparison is even more favorable, if we consider the other variants of Baldur, which achieve a proof rate of up to 47.9\%.

We observe that the comparison depends on the computational cost that we spend during inference.
While comparing the cost required for the two methods is involved, one measure we can use is the amount of computational resources reserved during proof generation. For a single proof, the language model approach using search~\cite{Jiang2022Thor} requires a TPUv3 with 8 cores for 216 seconds,\footnote{Section 4.1 in \cite{Jiang2022Thor} states that 1000 problems take around 60 TPU hours.} while our methodology also requires a TPUv3 with 8 cores for around 35 seconds to sample 64 proofs -- a difference of factor 6.
This argument disregards the time spent on proof checking, which is intentional: proof checking is done on CPUs, which is cheap compared to time spent on TPUs.
So, disentangling these two workloads can lead to significant reductions in computational cost.


\begin{tcolorbox}
\textbf{RA1}: These results demonstrate that LLMs can generate full proofs just as well as smaller language models augmented with a search strategy.
\end{tcolorbox}


\begin{figure*}
    \centering
    \begin{tikzpicture}
    \begin{axis}[ 
    width=12cm,
    height=7cm,
    legend pos=south east,
    xlabel={number of proof attempts},
    ylabel={ratio of proven theorems},
    ]
    \addplot table [x=inference cost, y=ratio, col sep=comma] {generate.csv};
    \addplot table [x=inference cost, y=ratio, col sep=comma] {generateandrepair.csv};
    \addplot table [x=inference cost, y=ratio, col sep=comma] {generateandrepairblind.csv};
    \addlegendentry{Generate}
    \addlegendentry{Generate+Repair}
    \addlegendentry{Generate+Repair (no err msg)}
    \end{axis}
    \end{tikzpicture}
    \caption{Ratio of theorems proven vs inference cost.}
    \label{fig:blueplot}
\end{figure*}
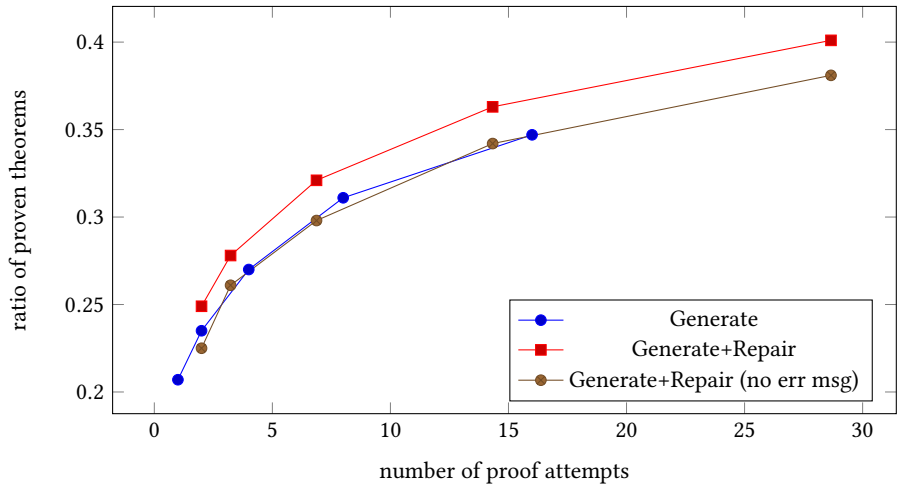

\subsection{RQ2: Can LLMs be used to repair proofs?}
\label{ssec:rq2}

We trained models for proof generation and repair as detailed in
Section~\ref{sec:method}.
If we sample from the proof generation model once with temperature 0, collect the failed proofs, and then repair once with temperature 0, we generate an additional 266 or 4.2\% correct proofs.
However, in this comparison, the generate + repair approach uses two samples, while the generate approach has only one sample.
For a fair comparison, we have to compare the repair approach to the generate approach with additional inference attempts.

In Figure~\ref{fig:blueplot}, we plot the proof success rate of the generate approach and the repair approach against the number of proof attempts.
Note that the number of samples for the repair approach does not perfectly align with the number of samples for the generate approach.
This is because the generate approach tends to produce multiple copies of the same proofs, which we deduplicate before repair, and only generate one repair attempt per failed proof attempt.
For each of the number of samples of the generate approach, we tune the temperature in the range of 0.0 to 1.4 in increments of 0.2, and we always use temperature 0 for the repair approach.

We observe that the repair approach consistently outperforms the plain proof generation model, which only uses the theorem statement as input.
However, this does not yet answer the question of where those gains from.
To shed some light on this question, we trained another repair model that is given the same information, except that it does not see the error message.
Plotting the proof success rate of this model in Figure~\ref{fig:blueplot} shows us that while it is able to prove additional theorems, it does not surpass the performance of the generate model when normalized for inference cost.
This suggests that the information in the error message is crucial for the observed gains of the repair approach.

\begin{tcolorbox}
\textbf{RA2}: LLMs can be used to repair proofs, including their own failed proof attempts,
and this can boost overall proving power.
\end{tcolorbox}

\subsection{RQ3: Can LLMs benefit from using the context of the theorem?}
\label{ssec:rq3}

In Table~\ref{tab:results}, we report the impact of adding theory file context to our plain generation approach.
At 64 samples, the proof rate increases from 40.7\% to 47.5\% for the same model size.
In Figure~\ref{fig:sampling}, we plot the proof success rate of the 
generation model with and without context against the number of proof attempts. We observe that the proof generation models with context consistently outperform the plain generation model. 

To get a better understanding of where these gains are coming from, we inspected 5 randomly sampled examples that the model using context was able to solve, but the plain generation model could not. Appendix~\ref{app:context_examples} displays these examples and further details the process we used to select them.

While the sample size is not large enough to make quantitative judgements, it appears that the model frequently makes use of similar proofs in the context.
We observe that for 3 of the 5 examples (see Appendices~\ref{app:ex1},~\ref{app:ex3},~\ref{app:ex5}) the model readily {\bf copies and adapts} proofs that exist in its context. For another example (see Appendix~\ref{app:ex2}), the model made use of a premise that did not occur in its context, which happened to also be used in the ground truth proof, but with a different tactic. In the final example (see Appendix~\ref{app:ex4}), the model found a simpler proof that did not occur like this in the context.
This suggests that the addition of context does not play the same role as premise selection. 




\begin{tcolorbox}
\textbf{RA3}: LLMs can benefit from the context in which the theorem occurred in the theory file, both quantitatively by increasing proving power, and qualitatively by copying and adapting nearby proofs.
\end{tcolorbox}

\subsection{RQ4: Does the size of the LLM affect proof synthesis effectiveness?}
\label{ssec:rq4}
We fine-tuned and evaluated the 62b version of Minerva on the proof generation task with context. In Table~\ref{tab:results}, we report that for 16 samples, the large model can prove an additional 1.3\% over the 8b model, resulting in a total proof rate of 42.2\%. For 64 samples, the large model can prove an additional 0.4\% over the 8b model, resulting in a total proof rate of 47.9\%.

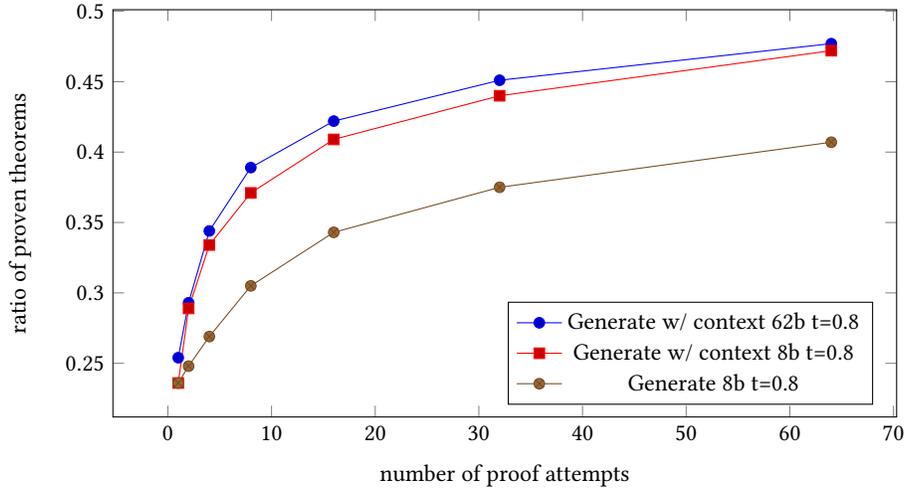
\begin{figure*}
    \centering
    \begin{tikzpicture}
    \begin{axis}[
    width=12cm,
    height=7cm,
    legend pos=south east,
    xlabel={number of proof attempts},
    ylabel={ratio of proven theorems},
    ]
    \addplot table [x=inference cost, y=ratio, col sep=comma] {62bcontext_64samples_t0.8.csv};
    \addlegendentry{Generate w/ context 62b t=0.8}
    \addplot table [x=inference cost, y=ratio, col sep=comma]{8bgenerate_context.t0.8.csv};
    \addlegendentry{Generate w/ context 8b t=0.8}
    \addplot table [x=inference cost, y=ratio, col sep=comma] {8bgenerate.t0.8.csv};
    \addlegendentry{Generate 8b t=0.8}
    \end{axis}
    \end{tikzpicture}
    \caption{Ratio of theorems proven vs inference cost for models with different sizes and temperatures.}
    \label{fig:sampling}
\end{figure*}

In Figure~\ref{fig:sampling}, we plot the proof success rate of the generation model with context for the 8b model and the 62b model against the number of proof attempts. We observe that the 62b proof generation model with context outperforms the 8b proof generation model with context.
One caveat here is that we were not able to tune hyperparameters as well due to the higher cost of these experiments, so an optimally tuned 62b model may perform even better.

\begin{tcolorbox}
\textbf{RA4}: Theorem proving performance improves with the scale of the language model.
\end{tcolorbox}

\subsection{RQ5: How do LLMs compare to other SOTA proof generation methods?}
\label{ssec:rq5}

While comparisons across different neural theorem provers are hard in general, we can compare to Thor~\cite{Jiang2022Thor}, one of the strongest approaches available.
Thor also relies on language models, but uses smaller models (700m parameters) and uses a different kind of proof step as its prediction target. Instead of using the human ground truth proofs, Thor generates a new training set and aims to solve each proof step by generating a declarative statement, which is then solved using Sledgehammer.
That is, Thor disentangles the planning stage of the next proof step, which is the specification of the target state (using a ``have'' statement) and premise selection, which is done by Sledgehammer.
This enables Thor to solve a total of 57\% of the problems.

In contrast, our approach solves up to 47.9\% of the problems.
While there is a significant gap, we argue that the means by which the two techniques improve over plain language modeling are largely orthogonal.
In Table~\ref{tab:results}, we report a large gain from 57\% to 65.7\% when we consider the union of \sysname and Thor, which supports this hypothesis.

\begin{table}

    \centering
    \begin{tabular}{lccc}
        \hline
        AFP Topic & Test set & \sysname & Thor \\
        \hline
         Computer Science &  4,019 & 50.0\% & 57.5\%\\
         Logic & \phantom{0,}966 & 51.6\% & 53.6\%\\
         Mathematics & 2,200 & 41.9\% & 50.5\%\\
         Tools & \phantom{0,}102 & 53.9\% & 51.8\%\\ 
         \hline
    \end{tabular}
    \caption{Proof rate by AFP topic classification, and the number of theorems in each category. While there are only 6336 theorems in total in the test set, the projects these theorems appear in can fall into multiple topics.}
    \label{tab:afp_topics}
\end{table}

We compare the proof rate of \sysname and Thor on different types of problems. The AFP is indexed by topic and there are four overarching topics: computer science, logic, 
mathematics, and tools. The authors of individual proof developments self-identity which topics
their projects fall into. We use these provided topic labels to determine the categories of problems from our test that \sysname and Thor can most effectively solve. 
Table~\ref{tab:afp_topics} shows the breakdown of which theorems in the
test set fall into which topics and \sysname's and Thor's proof success rates on these theorems.
In terms of relative performance, \sysname performs better than Thor on problems related to tools and similarly on problems related to logic. 
We observe that \sysname's performance on mathematics and computer science is less than that of Thor's performance. For mathematics proofs, we hypothesize that premise selection may be particularly useful, and Thor's use of Sledgehammer is likely what gives it a leg up on solving these mathematics problems.

\begin{tcolorbox}
\textbf{RA5}: Our findings suggest that LLM-based methods and search-based methods are complementary, and together can lead to large gains in proving power.
\end{tcolorbox}

\section{Discussion: What's Next?}
\label{ssec:discussion}

Our evaluation shows that LLMs can generate whole proofs at once, and can repair their own mistakes,
forming the basis for an effective and simple approach to proof synthesis.
Moving forward, we find three directions particularly promising:

\begin{enumerate}[labelwidth=0.7em, labelsep=0.6em, topsep=-0.5ex, itemsep=0ex, parsep=0ex, leftmargin=2em]

    \item integrating proof generation and proof repair models into a new \textbf{learnable proof search} strategy,

    \item investigating \textbf{alternative data splits} corresponding to different goals, and

    \item evaluating these techniques across \textbf{different proof assistants}.
\end{enumerate}

\paragraph{Learnable Proof Search}

While our generate + repair approach to proof synthesis lets us avoid costly proof search
procedures, it also lends itself to a new proof search strategy.
The search strategy would work as follows:

\begin{enumerate}[labelwidth=0.7em, labelsep=0.6em, topsep=-0.5ex, itemsep=0ex, parsep=0ex, leftmargin=2em]

    \item use the generation model to sample candidate proofs,

    \item use the repair model to attempt to repair those proofs, and

    \item continue to use the repair model to repair the repair-model-generated attempts from (2).

\end{enumerate}
This paves the way for a learnable proof search strategy.

We demonstrate a proof-of-concept of this new proof search 
strategy. We sample once using the generation model, repair the 
generated sample using the repair model, and
repair the repair model's attempt using the repair model. When 
using both models, we sample with temperature 0. So the inference
cost in this setup is 3 (1 for the first generation, 1 for the first repair, and 1 for the second repair).

The generate + repair approach with inference cost of 2 
proves 24.9\% of the test set theorems. With a second repair attempt, it proves
an additional 1.3\%, for a total of 26.2\%. The generation
approach with inference cost of 3 proves 25.4\%, which is 0.8\% less than the second repair attempt for the same inference cost.

To make this a more viable proof search strategy, future work needs to 
focus on generating proof repair training data that better mirrors the required changes 
for the subsequent repair attempts. When proof checking, the resulting error message is 
for the first occurring error, typically from the first couple of lines of 
the predicted proof. So the proof repair model will only 
learn to address these types of errors. An alternative approach could be, for example, 
to take the training examples from the proof generation model and use the first few lines of the human-written ground truth proof as a \emph{proof prefix}. We could then 
concatenate this proof prefix to the end of the input. Since it is a decoder-only model, 
we can simply sample the model's attempt at the rest of the proof. If the proof prefix 
concatenated with the rest of the proof does not check, then that can serve as a new
training example for the proof repair model.

\paragraph{Alternative Data Splits}

The PISA benchmark that we use to evaluate our approach commits to a particular
data split between training data and testing data.
It is interesting to note, however, that different data splits may themselves correspond to
different goals, even fixing the same evaluation task and metric.
Moving forward, it may be useful to consider different kinds of data splits
corresponding to different goals, even fixing the same dataset and benchmark suite.
Here, we consider two different splits: \emph{theorem-wise} and \emph{project-wise}.

PISA uses a random theorem-wise split of the theorems appearing the AFP. 
This means that for any theorem in the test set, the theorems and (the corresponding proofs) that appear 
before or after that theorem may be in the training set. This split is useful to evaluate since a forward-looking goal of neural
theorem prover researchers is to integrate these tools directly into proof assistants, where
they could make use of the full project context.
That project context may include human-written proofs of nearby theorems that look
similar (or even identical) to one another --- automatically repurposing and adapting those proofs
can be quite fruitful.

By contrast with PISA, CoqGym~\cite{yang2019learning}, the neural theorem prover benchmark suite for the Coq proof
assistant, uses a project-wise split, where training and testing data come from entirely different projects.
This is useful when the goal is to help proof engineers who start completely new projects
and want an automated proof synthesis tool to prove as much as it can.
A tool that is trained and evaluated in a setting where it expects that it has seen proofs in a 
given proof development, as may happen with a theorem-wise split,
may not perform as well in this new setting. 
Explicit consideration for the data split and the
goals it achieves may help drive neural theorem proving research even further.

\paragraph{Different Proof Assistants}

To make better sense of new strides in neural theorem proving,
it makes sense to evaluate the same techniques across many different proof assistants.
But this remains challenging.
Consider once again the problem of data splits:
since prover developments that evaluate on CoqGym~\cite{First20oopsla, First22icse} follow the same project-wise split as CoqGym, it can be hard to make sense of how those developments
compare to those trained and evaluated using theorem-wise data splits, like our own \sysname.

We used an established benchmark of Isabelle/HOL proofs to fairly compare \sysname to prior work and to increase the chances that our results generalize.  However, we observed that search-based proof-synthesis tools for other proof assistants tend to prove a smaller fraction of theorems than we have found in our work.  For example, Diva~\cite{First22icse}, the current state of the art for the Coq proof assistant, proves 33.8\% of its benchmark automatically.  This could be a reflection of size and quality of the available training data or the complexity of the available evaluation data (which, by necessity, is different from what we use because it involves theorems and proofs in different languages), or a more fundamental difference in the complexity of synthesizing proofs in these respective languages.  

Future work should allow for direct comparisons by porting the developed techniques across proof assistants. Cross-proof-assistant benchmark suites may help substantially with this, but
still have their limitations. For example, MiniF2F~\cite{zhengminif2f} implements the same benchmark suite for Math Olympiad problems across many different proof assistants. But math problems are not evenly represented across proof assistants, which draw different user communities with different emphases. 
Fair comparisons between proof assistants are hard, but we do believe they are necessary.

\section{Related Work}
\label{sec:related}


Existing methods for automating formal theorem proving can be classified into two categories, hammers and search-based methods.  Hammers, such as CoqHammer~\cite{Czajka18} and Sledgehammer~\cite{Paulson23}, iteratively use a set of
precomputed mathematical facts to attempt to ``hammer'' out a proof. While hammers are powerful, they lack the ability to employ certain tactics, such as induction, preventing them from proving certain large classes of theorems.  Search-based methods use a prediction model that, given some information about a partially written proof, the target theorem being proven, and the current proof state, predicts a set of next likely proof steps.  The methods then use metaheuristic search~\cite{Harman07} to attempt to synthesize a proof.  They iterate querying the prediction model for the likely next steps and using the proof assistant to get feedback on those steps and prune non-promising paths, generating a search tree of possible proofs. The proof assistant also determines when the proof is complete.  The tools mostly differ in the prediction model they use, which are typically learned automatically.  For example, ASTactic uses only the proof state~\cite{yang2019learning}, TacTok uses the proof state and the partially written proof script~\cite{First20oopsla}, Diva (which combines the use of many models) also uses the proof term~\cite{first2022diva}, and Passport also uses identifier information~\cite{Sanchez-Stern22passport}.
Other search-based techniques include Tactician~\cite{tactician}, 
Proverbot9001~\cite{proverbot9001}, and GamePad~\cite{Huang2019Gamepad} for Coq;
TacticToe~\cite{gauthier2021tactictoe} for HOL4;
and DeepHOL~\cite{Bansal2019HOList, Paliwal2020Graph} for HOL Light.
Prior work has found that hammers and search-based methods are complementary, each often proving theorems the other cannot~\cite{yang2019learning, First20oopsla, first2022diva}.
Thor \cite{Jiang2022Thor} combines a search-based method with a hammer, using both a prediction model and Sledgehammer in its search. In contrast, our approach uses an LLM to generate an entire proof at once, and then to one-shot repair it.

The most closely related work to ours is LISA~\citep{Jiang21Lisa}, which fine-tunes a pretrained language model on a large Isabelle/HOL proof corpus, and uses it inside of a search procedure to predict proof steps.
GPT-f~\citep{polu2020gptf} likewise combines a generative language model with proof search to target the Metamath proof language.
A Monte-Carlo tree search approach outperforms GPT-f in Lean~\cite{Lample2022HypertreeProofSearch}.

TacticZero~\citep{Wu2021TacticZero} learns not just tactics but also proof search strategies for end-to-end proof synthesis, rather than relying on a single fixed proof search strategy like other neural theorem proving approaches. 
The approach works by way of deep reinforcement learning, and improves over the previous state of the art on a benchmark for the HOL4 theorem prover.

A related problem to neural theorem proving is \emph{autoformalization}: the automatic translation
of natural language specifications and proofs into formal, machine-checkable specifications and proofs.
LLMs have shown promise for autoformalization of specifications,
and automatically generated proofs of the resulting autoformalized specifications have 
been used to improve a neural theorem prover on a widely used benchmark suite in Isabelle/HOL~\cite{Wu2022autoformalization}.
ProofNet~\cite{azerbayev2022ProofNet} introduces a dataset and benchmark suite for autoformalization in Lean, based on undergraduate mathematics, and shows preliminary promising results
autoformalizing proofs on that benchmark using Codex~\cite{chen2021codex} with few-short learning.
Autoformalization of both theorems and proofs in Coq shows 
promise on a small preliminary benchmark suite~\cite{garett2023}.
Autoformalization for specification logics in verification
is also promising~\cite{Hahn2022LTLAutoformalization}.

The Draft, Sketch, and Prove method (DSP)~\cite{jiang2022draftsketchprove} presents a hybrid between theorem proving and autoformalization, which, similar to our approach, makes use of LLMs for theorem proving.
It provides informal proofs as drafts for the LLM to translate into a formal proof sketch, which is then proven via Sledgehammer.
In contrast, we use fine-tuning for LLMs, do not make use of Sledgehammer, and do not rely on the availability of natural language proofs.


Pretrained language models can be used to answer natural-language mathematics questions~\cite{Noorbakhsh21}.
Large language models, such as Minerva~\cite{Lewkowycz2022Minerva} and PaLM~\cite{chowdhery2022palm}, have been evaluated on natural language mathematics benchmarks, such as GSM8k~\cite{Cobbe2021TrainingVerifiers} and MATH~\cite{hendrycks2021math}.
The ProofNet~\citep{azerbayev2022ProofNet} benchmark suite mentioned above includes informal proofs
alongside formal proofs as a benchmark.

We introduce the proof repair task, with error messages. This is a new machine learning task for formal proofs.  We show that solving this task improves neural theorem proving performance.
Proof engineers perform proof repair constantly during formal proof development~\citep{ringer2020replica}.
Automating this task first arose with the advent of
symbolic tools for automatic proof repair in the Coq proof assistant~\citep{ringer2021proof},
and has since made its way into tools for other proof systems~\citep{masci2022proof}.
Our work is among the first to explore proof repair in a machine learning context, and 
the first we are aware of to use error messages for a proof repair task,
and to use repair to improve performance of proof synthesis.

There are numerous other tasks that machine learning tools
for proofs consider that may either help users with proof development directly, or improve neural theorem proving performance themselves.
For example, PaMpeR~\citep{pamper} predicts proof methods alongside explanations in Isabelle/HOL.
ACL2(ml)~\citep{acl2ml} generates helper lemmas and suggests similar theorems in ACL2.
Other popular tasks leveraging machine learning include premise selection and datatype alignment, and are described in more detail in QED at Large~\citep{ringer2019qed}.

Our approach can help minimize human effort in formal verification by automatically synthesizing proofs for some theorems.  
Other tools that assist humans writing formal verification proofs can similarly save time, and can be complementary to our work for theorems \sysname cannot prove fully automatically.
iCoq~\cite{Celik17, Celik18}, and its parallelized version
PiCoq~\cite{Palmskog18}, find failing proof scripts in evolving projects by
prioritizing proof scripts affected by a revision. iCoq tracks fine-grained
dependencies between Coq definitions, propositions, and proof scripts to
narrow down the potentially affected proof scripts. QuickChick~\cite{Lampropoulos17}, a random testing
tool for Coq, searches for counterexamples to executable theorems, helping a
programmer to become more confident that a theorem is correct.
Roosterize~\cite{Nie21, Nie20ijcar} can suggest names for lemmas, and language models can also help automatically format proofs~\cite{Nie20}, both improving readability and maintainability.  
Mutation analysis can identify weak specifications, when
mutating definitions does not break their proofs~\cite{Celik19, Jain20}. The
mutation operators could, hypothetically, be applied in repair and in
providing feedback for developers as to why a proof has broken.


The automated program repair field studies the task of taking a program with a bug, evidenced by one or more failing tests, and automatically producing a modified version of the program that passes all the tests~\cite{LeGoues19}.  
Generate-and-validate repair techniques use search-based techniques or predefined templates to generate many syntactic candidate patches,
validating them against the tests (e.g., GenProg~\cite{LeGoues12b},
Prophet~\cite{Long16}, AE~\cite{Weimer13}, HDRepair~\cite{Le16},
ErrDoc~\cite{Tian17}, JAID~\cite{Chen17}, Qlose~\cite{DAntoni16}, and
Par~\cite{Kim13}, 
ssFix~\cite{Xin17},
CapGen~\cite{Wen18},
SimFix~\cite{Jiang18},
Hercules~\cite{Saha19},
Recoder~\cite{Zhu21},
among others). Techniques such as DeepFix~\cite{Gupta17}
and ELIXIR~\cite{Saha17} use learned models to predict erroneous program locations, as well as the patches.
It is possible to learn how to repair errors together by learning how to create errors, which can increase the amount of available training data, but poses an additional challenge of learning to approximate making human-like errors~\cite{yasunaga2021break}.  
Unfortunately, these automated program repair techniques often overfit to the available tests and produce patches that, while passing all the tests, fail to encode the developers' intent~\cite{Smith15fse, Motwani22, Noda20, Qi15}. 
Improving the quality of the resulting repairs can be done via improving fault localization
strategies~\cite{Motwani23icse, Assiri17, Yang18, Sun18, Koyuncu19, Jiang19, Lou20}, 
patch generation algorithms (e.g., heuristic-based~\cite{LeGoues12b, Long16, Tian17, Wen18, Jiang18, Petke20}, 
constraint-based~\cite{Afzal21, Ke15ase, Wang18, Gulwani18, Mechtaev18}, 
and learning-based~\cite{Chen19, Gupta17, Saha17}), and 
patch validation methodologies~\cite{Wang20, Yang17fse, Yu19, Tian20, Ye21}.
By contrast, in \sysname's domain of theorem proving, it is impossible to produce a proof that appears to prove the theorems, but actually fails to do so, because the theorem prover acts as an absolute oracle for the correctness of the proof.  As a result, it may be more difficult to produce a proof in the first place, but if techniques in this domain do produce proofs, they are guaranteed to be correct.

\section{Contributions}
\label{sec:Contributions}

This paper is the first to fine-tune large language models to generate entire proofs of theorems without the need for proof search or hammers.  We demonstrate that this approach is more effective and more efficient than prior methods that use one-step-at-a-time search-based generation, and that it is complementary to existing search-based and hammer-based approaches: Together, our \sysname and prior tools can fully automatically synthesize proofs for 65.7\% of the theorems in a large Isabelle/HOL benchmark, establishing a new state of the art. We further demonstrate that generate-and-repair improves proof synthesis when the language model is given access to the error messages produced by erroneous proofs.  

This work opens new avenues of research into (1)~using LLMs to automate theorem proving and simplify formal verification of software properties, (2)~repair approaches, both for proofs and, potentially, more traditional automated program repair tasks, and (3)~the use of context (e.g., failed synthesis attempts and error messages) in proof generation.  Our very encouraging results suggest a bright future for automated proof generation and repair using LLMs.

\section*{Acknowledgments}
We thank Stella Biderman, Ernest Davis, and others who provided feedback on an earlier draft of this paper.
This work is supported by the Defense Advanced Research Projects Agency under
grant no.\ DARPA HR0011-22-9-0063, and by the National Science Foundation
under grant no.\ CCF-2210243.

\balance 
\bibliography{repairPaper}
\bibliographystyle{ACM-Reference-Format}

\appendix





\section{Examples of Proof Generation with Context}
\label{app:context_examples}

We provide a number of examples that the model using context could solve but the plain proof generation model could not.
We determined the lists of problems each model could solve, computed their difference, and then sampled 5 examples uniformly at random.
For examples that had multiple correct proofs generated by the model, we selected one at random.
We modified whitespace in the examples to make them more readable with the reduced line length.
Further, we truncated the examples on the left to help with readability, but we inspected also the full context to ensure that our conclusions below are not affected.
Each example consists of the ``context and problem statement'', the ``ground truth proof'', and the ``generated proof''.

We can observe in examples 1, 3, and 5 that the model readily {\bf copies and adapts} proofs that exist in its context.
In example 2, the model made use of a premise that did not occur in its context, which happened to also be used by the ground truth proof, but with a different tactic. In example 4, the model found a simpler proof that did not occur like this in the context.

\subsection{Example 1}
\label{app:ex1}

\vspace{10pt}
Context and problem statement:
\begin{lstlisting}
lemma (in Interpretation) InterpExprWellDefined:
  "L\<lbrakk>Vx : A \<turnstile> e : 
    B\<rbrakk> \<rightarrow> i \<Longrightarrow> 
      Sig iS \<triangleright> Vx : 
        A  \<turnstile> e : B"
  apply (rule Interp.cases)
  by auto

lemma (in Interpretation) WellDefined:
  "L\<lbrakk>\<phi>\<rbrakk> \<rightarrow> i
    \<Longrightarrow> Sig iS \<triangleright> \<phi>"
  apply(rule Interp.cases)
  by (auto simp add: InterpExprWellDefined)

lemma (in Interpretation) Bool:
  "L\<lbrakk>\<phi>\<rbrakk> \<rightarrow>
    (IBool i) \<Longrightarrow> \<exists> A B e d .
      \<phi> = (Vx : A \<turnstile> e \<equiv> d : B)"
\end{lstlisting}

\vspace{10pt}
\noindent
Ground truth proof:
\begin{lstlisting}
apply(rule Interp.cases)
by auto
\end{lstlisting}

\vspace{10pt}
\noindent
Generated proof:
\begin{lstlisting}
apply (rule Interp.cases)
by (auto simp add: InterpExprWellDefined)
\end{lstlisting}

\subsection{Example 2}
\label{app:ex2}
\vspace{10pt}
Context and problem statement:
\begin{lstlisting}
lemma Aeta_expand_0[simp]:
  "Aeta_expand 0 e = e"
  by transfer simp

lemma Aeta_expand_inc[simp]:
  "Aeta_expand (inc\<cdot>n) e = 
    (Lam [fresh_var e].
      Aeta_expand n (App e (fresh_var e)))"
  apply (simp add: inc_def)
  by transfer simp

lemma subst_Aeta_expand:
  "(Aeta_expand n e)[x::=y] =
    Aeta_expand n e[x::=y]"
\end{lstlisting}

\vspace{10pt}
\noindent
Ground truth proof:
\begin{lstlisting}
by transfer (rule subst_eta_expand)
\end{lstlisting}

\vspace{10pt}
\noindent
Generated proof:
\begin{lstlisting}
by transfer (simp add: subst_eta_expand)
\end{lstlisting}

\subsection{Example 3}
\label{app:ex3}
\vspace{10pt}
\noindent
Context and problem statement:
\begin{lstlisting}
subsubsection \<open>Store and Assignment\<close>

type_synonym \'a store = "string  \<Rightarrow> \'a"

definition gets ::
  "string \<Rightarrow> (\'a store \<Rightarrow> \'a) 
    \<Rightarrow> \'a store rel" ("_ ::= _" [70, 65] 61)
  where "v ::= e = {(s, s(v := e s)) |s. True}"

lemma H_assign:
  "rel_kat.H \<lceil>\<lambda>s. 
    P (s (v := e s))\<rceil> (v ::= e) \<lceil>P\<rceil>"
  by (auto simp: 
    gets_def
    rel_kat.H_def
    rel_kat.t_op_def
    rel_at_def)

lemma H_assign_var:
  "(\<forall>s. P s \<longrightarrow> Q (s (v := e s)))
    \<Longrightarrow>
    rel_kat.H 
      \<lceil>P\<rceil> (v ::= e) \<lceil>Q\<rceil>"
\end{lstlisting}

\vspace{10pt}
\noindent
Ground truth proof:
\begin{lstlisting}
by (auto simp:
  gets_def
  rel_kat.H_def
  rel_kat.t_op_def
  rel_at_def)
\end{lstlisting}

\vspace{10pt}
\noindent
Generated proof:
\begin{lstlisting}
by (auto simp:
  gets_def
  rel_kat.H_def
  rel_kat.t_op_def
  rel_at_def)
\end{lstlisting}

\subsection{Example 4}
\label{app:ex4}
\vspace{10pt}
\noindent
Context and problem statement:
\begin{lstlisting}
section \<open>Primitive words\<close>

text\<open>If a word $w$ is not a non-trivial power of
some other word, we say it is primitive.\<close>

definition primitive :: "\'a list \<Rightarrow> bool"
  where "primitive u = (\<forall> r k. 
      r\<^sup>@k = u \<longrightarrow> k = 1)"

lemma primI[intro]:
  "(\<And> r k. r\<^sup>@k = u \<Longrightarrow> k = 1) 
  \<Longrightarrow> primitive u"
  by (simp add: primitive_def)

lemma prim_nemp:
  "primitive u \<Longrightarrow> u \<noteq> \<epsilon>"
\end{lstlisting}

\vspace{10pt}
\noindent
Ground truth proof:
\begin{lstlisting}
proof-
have "u = \<epsilon> \<Longrightarrow>
  \<epsilon>\<^sup>@0 = u"
by simp
thus "primitive u \<Longrightarrow>
  u \<noteq> \<epsilon>"
using primitive_def zero_neq_one
by blast
qed
\end{lstlisting}

\vspace{10pt}
\noindent
Generated proof:
\begin{lstlisting}
using primitive_def
by force
\end{lstlisting}

\subsection{Example 5}
\label{app:ex5}
\vspace{10pt}
\noindent
Context and problem statement:
\begin{lstlisting}
lemma drop_take_drop:
  "drop n (take m xs) @ drop m xs = drop (min n m) xs"
proof cases
  assume A: "\<not>m < n"
  then show ?thesis
    using drop_append[of n "take m xs" "drop m xs"]
  by (cases "length xs < n")
       (auto simp: not_less min_def)
qed (auto simp: min_def)

lemma not_distinct_decomp_min_prefix:
  assumes "\<not> distinct ws"
  shows "\<exists> xs ys zs y.
    ws = xs @ y # ys @ y # zs \<and>
    distinct xs \<and> 
    y \<notin> set xs \<and> 
    y \<notin> set ys "
proof -
  obtain xs y ys where 
    "y \<in> set xs" "distinct xs" "ws = xs @ y # ys"
    using assms by (auto simp: not_distinct_conv_prefix)
  moreover then obtain xs\' ys\' 
    where "xs = xs\' @ y # ys\'" 
    by (auto simp: in_set_conv_decomp)
  ultimately show ?thesis by auto
qed

lemma not_distinct_decomp_min_not_distinct:
  assumes "\<not> distinct ws"
  shows "\<exists>xs y ys zs. 
    ws = xs @ y # ys @ y # zs \<and>
    distinct (ys @ [y])"
\end{lstlisting}

\vspace{10pt}
\noindent
Ground truth proof:
\begin{lstlisting}
using assms
proof (induct ws)
  case (Cons w ws)
  show ?case
  proof (cases "distinct ws")
    case True
    then obtain xs ys where
        "ws = xs @ w # ys" "w \<notin> set xs"
      using Cons.prems
        by (fastforce dest: split_list_first)
    then have "distinct 
        (xs @ [w])" "w # ws = [] @ w # xs @ w # ys"
      using \<open>distinct ws\<close> by auto
    then show ?thesis by blast
  next
    case False
    then obtain xs y ys zs
        where "ws = xs @ y # ys @ y # zs \<and>
               distinct (ys @ [y])"
      using Cons by auto
    then have "w # ws = (w # xs) @ y # ys @ y # zs \<and>
               distinct (ys @ [y])"
      by simp
    then show ?thesis by blast
  qed
qed simp
\end{lstlisting}

\vspace{10pt}
\noindent
Generated proof:
\begin{lstlisting}
proof -
  obtain xs y ys where
    "y \<in> set xs" "distinct xs" "ws = xs @ y # ys"
  using assms by (auto simp: not_distinct_conv_prefix)
  moreover then obtain xs' ys'
    where "xs = xs' @ y # ys'"
    by (auto simp: in_set_conv_decomp)
  ultimately
  show ?thesis by auto
qed
\end{lstlisting}





\end{document}